\documentclass[conference]{IEEEtran}
\IEEEoverridecommandlockouts
\usepackage{amsmath,amssymb,amsfonts}
\usepackage{graphicx}
\usepackage{textcomp}
\usepackage[hidelinks]{hyperref}
\usepackage{cleveref}
\usepackage{xcolor}
\def\BibTeX{{\rm B\kern-.05em{\sc i\kern-.025em b}\kern-.08em
    T\kern-.1667em\lower.7ex\hbox{E}\kern-.125emX}}

\usepackage{custom_def}

\begin{document}

\title{Structured Sampling for \\Robust Euclidean Distance Geometry\\
\thanks{This work is partially supported by NSF DMS 2208392 and 2304489.
}
}

\author{\IEEEauthorblockN{Chandra Kundu}
\IEEEauthorblockA{\textit{Department of Statistics and Data Science}\\
\textit{University of Central Florida}\\
Orlando, FL 32816, USA\\
\href{mailto:chandra.kundu@ucf.edu}{chandra.kundu@ucf.edu}
}
\and
\IEEEauthorblockN{Abiy Tasissa}
\IEEEauthorblockA{\textit{Department of Mathematics}\\
\textit{Tufts University}\\
Medford, MA 02155, USA\\
\href{mailto:abiy.tasissa@tufts.edu}{abiy.tasissa@tufts.edu}
}
\and
\IEEEauthorblockN{HanQin Cai}
\IEEEauthorblockA{\textit{Department of Statistics and Data Science}\\
\textit{Department of Computer Science} \\
\textit{University of Central Florida}\\
Orlando, FL 32816, USA\\
\href{mailto:hqcai@ucf.edu}{hqcai@ucf.edu}
}
}

\maketitle

\begin{abstract}
This paper addresses the problem of estimating the positions of points from distance measurements corrupted by sparse outliers. Specifically, we consider a setting with two types of nodes: anchor nodes, for which exact distances to each other are known, and target nodes, for which complete but corrupted distance measurements to the anchors are available. To tackle this problem, we propose a novel algorithm powered by Nyström method and robust principal component analysis. Our method is computationally efficient as it processes only a localized subset of the distance matrix and does not require distance measurements between target nodes. Empirical evaluations on synthetic datasets, designed to mimic sensor localization, and on molecular experiments, demonstrate that our algorithm achieves accurate recovery with a modest number of anchors, even in the presence of high levels of sparse outliers.
\end{abstract}

\begin{IEEEkeywords}
Euclidean distance geometry, outlier removal, Nyström approximation, sensor localization, protein structure
\end{IEEEkeywords}

\section{Introduction}

Distance matrices naturally arise in numerous applied science problems, serving as a powerful quantitative tool to capture the similarity or dissimilarity between underlying entities \cite{mikolov2013efficient,yang2018learned}. Given a distance matrix, a common task is to determine a set of points in a low-dimensional space that approximately or exactly realize the given distances. Such embeddings are valuable for visualization and enable further inference. When the matrix represents a complete set of exact Euclidean distances, computed using the Euclidean norm, a standard technique to obtain the embedding is the classical multidimensional scaling (MDS) algorithm \cite{young1938discussion,torgerson1952multidimensional,torgerson1958theory,gower1966some}. However, in many real-world applications, such as protein structure prediction and sensor localization, distance measurements are often subject to noise and outliers \cite{havel2002distance,xiao2013robust}. These imperfections necessitate stability analysis and robust algorithms \cite{arias2020perturbation,sibson1979studies}. 

In addition to noise and outliers, missing distances also pose another significant challenge. Factors such as geography, climate, device precision, or the cost of acquiring measurements often result in incomplete distance matrices \cite{aldibaja2016improving,marti2013multi,clore1993exploring}. In such cases, classical MDS cannot be directly applied. The task of finding an underlying set of points from an incomplete distance matrix is referred to as the Euclidean Distance Geometry (EDG) problem. For this problem, various theoretical and algorithmic solutions tailored to specific applications and/or models of missing distances have been developed \cite{fang2013using,lavor2012discretizable, abiy_exact,Nguyen2019,Smith2024, ghosh2024sampleefficientgeometryreconstructioneuclidean}.

In certain scenarios, missing distances exhibit a structured pattern. For instance, in the sensor localization problem  \cite{mao2007wireless,kuriakose2014review}, there are typically two types of nodes: anchor nodes with known locations and target nodes with unknown locations. A standard setup involves knowing the distances between the target nodes and all anchors while lacking information about distances between the target nodes themselves. Motivated by such setups, a recent work in \cite{lichtenberg2024localization} has developed a theoretical framework for recovering the locations of sensor nodes when provided with partial exact distance information between anchors and partial exact distance information between target nodes and anchors. In this paper, we consider a sampling model inspired by the standard sensor localization problem. Specifically, we address the scenario where distances between target nodes and anchors are corrupted by outliers, while distances between anchors are assumed to be exact. We propose a novel algorithm for this problem, powered by Nyström method and robust principal component analysis (RPCA). 

\subsection{Problem setup} \label{sec:problem setup}
Let $\p_1, \ldots, \p_T$ denote a set of $T$ points in $\mathbb{R}^r$. Define $\P = [\p_1, \ldots, \p_T] \in \mathbb{R}^{r \times T}$, a matrix whose columns represent these points. The squared Euclidean distance between two points $\p_i$ and $\p_j$ is given by $d_{i,j}^2 = \|\p_i - \p_j\|_2^2$, where $\|\cdot\|_2$ denotes the standard Euclidean norm. We can represent all the squared pairwise distances compactly using the squared distance matrix $\D\in \mathbb{R}^{T \times T}$, which is defined entrywise as $[\D]_{ij} = \|\p_i - \p_j\|_2^2$. 
Some algebraic manipulation yields the following relationship between $\P$ and $\D$:  
\begin{equation}
\label{eq:P_to_D}
\D = \mathbf{1} \, \text{diag}(\X)^\top + \text{diag}(\X) \, \mathbf{1}^\top - 2\X,
\end{equation}
where $\X = \P^\top \P$ is the Gram matrix,  $\mathbf{1}$ is a column vector of ones and $\text{diag}(\cdot)$ denotes a column vector of the diagonal entries of the matrix in consideration. Using linear-algebraic properties, it can be deduced that  
$\text{rank}(\D) \leq r + 2$ 
and $\text{rank}(\X) \le r$. In most real-world applications, such as sensor network localization and structure prediction, where $r = 3$ and $T \gg r$, both the squared distance matrix $\D$ and the Gram matrix $\X$ are low-rank. This low-rank property is particularly useful as it allows the application of well-established techniques from low-rank optimization.

In our setup, we consider two types of nodes: $m$ anchor nodes and $n$ target nodes. We are given all pairwise distances between the anchors, as well as the distances between the anchors and the target nodes. However, no distance information is available between the target nodes. Let $\p_1, \ldots, \p_m$ denote the positions of the $m$ anchor nodes, and $\p_{m+1}, \ldots, \p_T$ denote the positions of the $n$ target nodes, where $m + n = T$. Under this setup, the squared distance matrix $\D$ and the Gram matrix $\X$ can be expressed in the following block forms:  
\begin{gather}\label{eq:D}
   \D = \begin{bmatrix} 
  \E & \F \\
 \F^\top & \G 
\end{bmatrix},
\quad   \X = \begin{bmatrix} 
  \A & \B \\
 \B^\top & \C 
 \end{bmatrix}.
\end{gather}
In this paper, we study the problem of estimating the underlying set of points when we are given $\E$ exactly, $\F$ is prone to outliers and no distance information is available in $\G$. 




\subsection{Notation}
In this paper, we use the following notations. Uppercase boldface scripts, such as $\M$, denote matrices, while lowercase boldface scripts, such as $\v$, represent vectors. The $(i,j)$-th entry of a matrix $\M$ is denoted by $[\M]_{i,j}$. The transpose of a matrix $\M$ is denoted by $\M^\top$, and the pseudo-inverse of $\M$ is represented as $\M^\dagger$. The standard $\ell_2$-norm is denoted by $\|\cdot\|_2$. The vector of ones of size $m$ is denoted by $\1_m$, and the matrix of ones of size $m \times n$ is denoted by $\1_{m \times n}$. Finally, $\mathbb{E}([\M]_{i,j})$ denotes the mean of all elements of the matrix $\M$.


\section{Related work}
We give a brief overview of related work in Euclidean distance geometry, Nyström method and robust PCA. 

\subsection{Euclidean distance geometry and Nyström method}
The Euclidean Distance Geometry (EDG) problem focuses on recovering the configuration of underlying points from an incomplete distance matrix \cite{dokmanic2015euclidean,liberti2014euclidean}. Later studies show EDG can be posted as a generalized low-rank approximation problem with non-orthogonal measurements \cite{abiy_exact,Smith2024}. We focus on an anchor-based sampling model that has been established in \cite{cai2023matrix,lichtenberg2024localization}, however, they do not address the presence of outliers. Such a sampling model is closely related to Nyström approximation (a.k.a.~CUR approximation for asymmetric setting), which aims to obtain a low-rank matrix approximation by sampling a subset of rows and columns \cite{gittens2013revisiting,kumar2012sampling,mahoney2009cur,hamm2020perspectives,cai2021rcur,cai2024rccs}. However, the vanilla Nyström approximation is designed to sample and recover directly on the same low-rank space, i.e., based on orthogonal measurements. 
On the other hand, robustness against outliers has been studied in the sensor localization literature \cite{biswas2006semidefinite,ding2010sensor,fliege2019euclidean,sremac2019noisy,xiao2013robust,zhang2010outlier}; however, these methods are based either on orthogonal measurements or employing a semidefinite programming approach. 
To the best of our knowledge, this paper is the first work to study the robust EDG with non-orthogonal measurements. 

\subsection{Robust PCA}
Robust Principal Component Analysis (RPCA) is a fundamental machine learning tool for recovering a low-rank data matrix from sparse outlier corruptions. That is, given a corrupted observation $\Y=\L+\S$ where $\L$ is low-rank and $\S$ is sparse, RPCA recovers $\L$ and $\S$ simultaneously. RPCA is widely used in applications \cite{jang2016primary,gharibnezhad2015applying,luan2014extracting,chen2012clustering,gao2011robust,bouwmans2018applications,cai2021asap,cai2023hsgd,cai2024accelerating}, and numerous convex and non-convex solvers have been extensively studied under various settings \cite{wright2009robust, lin2010augmented,candes2011robust,cai2019accelerated,zhang2018robust,cai2021ircur,cai2021lrpca,tong2021accelerating,hamm2022RieCUR,giampouras2024guarantees,cai2025deeply}.




In this work, we employ RPCA to eliminate outliers from the $\F$ block of the distance matrix $\D$. Among various existing RPCA solvers, we use the non-convex \textit{Accelerated Alternating Projections} (AccAltProj) algorithm \cite{cai2019accelerated}, which is robust, lightweight, and highly efficient. AccAltProj has a recovery guarantee with linear convergence. Its main drawback is requiring the knowledge of the exact rank of $\L$; however, this is not an issue for EDG problems since the true rank is explicit as discussed in \Cref{sec:problem setup}.



\section{Background}

\subsection{Nyström approximation for distance matrices}
The squared distance matrix $\D$ has a block structure as shown in \eqref{eq:D}. To recover the blocks $\A$ and $\B$ of the Gram matrix $\X$ from $\D$, we leverage the relationship between $\X$ and $\D$ established in \citep{gower1982euclidean, gower1985properties}:
\begin{equation}
\label{eq:general_gram}
        \X = -\frac{1}{2}(\I - \bm{1}\s^\top)\D(\I - \bm{1}\s^\top), 
\end{equation}
where $\s$ is a vector with entries that sum to 1. This operation generates a family of valid solutions for $\X$ based on different choices of $\s$. Following \cite{platt2005fastmap}, we select $\s$ with entries equal to $1/m$ for the first $m$ positions and 0 elsewhere. In this paper, we propose a robust EDG framework that leverages non-orthogonal measurements to estimate the Gram matrix based on the relationship defined in \eqref{eq:general_gram}. The blocks $\A$ and $\B$ in \eqref{eq:general_gram} are computed as follows:
\begin{small}
\begin{align} 
    \A &= -\frac{1}{2}\left( \E - \frac{1}{m} \E \1_{m \times m} - \frac{1}{m} \1_{m \times m} \E + \mathbb{E}([\M]_{i,j}) \1_{m \times m} \right), \label{eq:findA}  \\
    \B &= -\frac{1}{2}\left( \F - \frac{1}{m} \1_{m \times m} \F - \frac{1}{m} \E \1_{m \times n} +  \mathbb{E}([\M]_{i,j}) \1_{m \times n} \right). \label{eq:findB}  
\end{align}
\end{small}%
However, we can only recover the blocks $\A$ and $\B$ of $\X$ from the $\E$ and $\F$ blocks of $\D$ and we still do not know $\C$. To recover the full Gram matrix $\X$, we apply the standard Nyström approximation procedure. This method approximates a positive semi-definite matrix as follows:
\begin{equation}
    \C \approx \B^\top \A^\dagger \B, \label{eq:nystrom}
\end{equation}
which becomes exact when $\rank(\A) = \rank(\X)$ \cite{kumar2009sampling}.

\section{Proposed approach}
Our goal is to reconstruct the Gram matrix $\X$ from a partially observed distance matrix $\D$, with available blocks $\E$ and $\F$, where $\F$ is corrupted by sparse noise. Since our method avoids dependency on the $\G$ block (distances between target nodes), this makes it more efficient and broadly applicable, as the $\G$ block might often be unavailable or incomplete due to the cost acquiring these distance measurements \cite{yang2024wireless, wang2006coverage}.

The reconstruction process begins by applying the RPCA algorithm to $\F$ to remove sparse noise, yielding an approximate $\hat{\F}$. By focusing only on the smaller $\F$ block rather than on the entire $\D$ matrix, our method reduces computational complexity. In addition, this localized approach is effective because $\F$ represents interactions between the anchor and target nodes, where sparse noise is likely to occur due to measurement errors or outliers, unlike $\G$, which might be typically missing or incomplete and, therefore, not directly involved in the reconstruction process.

Once the sparse noise is removed, we estimate the block $\B$ of the Gram matrix $\X$ from the clean $\hat{\F}$ and the available $\E$ block of $\D$. The reconstruction equations \eqref{eq:findA} and \eqref{eq:findB} are derived from the relationship between squared distance matrices and Gram matrices. The estimation of $\B$ relies only on the $\E$ and $\hat{\F}$ blocks, making the process efficient and independent of the missing $\G$ block. The Nyström approximation is then applied to estimate the missing $\C$ block of the Gram matrix $\X$, as described in \eqref{eq:nystrom}. Finally, the reconstructed blocks $\hat{\B}$, and $\C$ along with $\A$ are assembled to obtain the estimate of the Gram matrix $\X$. The complete reconstruction algorithm, namely Structured Robust EDG, is summarized in \Cref{algo:redg}.

\begin{algorithm}
\caption{Structured Robust EDG} \label{algo:redg}
\begin{algorithmic}[1]
    \State \textbf{Input:} $\E \in \Real^{m \times m}$: squared distance matrix among anchor nodes; $\F \in \Real^{n \times n}$: noisy squared distance matrix among anchor and target nodes; $\bar{r}=r+2$: rank of $\D$; $\RPCA$: the chosen RPCA solver. 
    \State Compute $\A$ from $\E$ using the formula in \eqref{eq:findA}.
    \State $[\hat{\F}, \hat{\S}] = \RPCA(\F, \bar{r})$.
    \State Compute $\hat{\B}$ from $\E$ and $\hat{\F}$ using the formula in \eqref{eq:findB}.
    \State $\hat{\C} = \hat{\B}^\top \A^\dagger \hat{\B}$.
    \State $\hat{\X} = \begin{bmatrix} 
    \A & \hat{\B} \\
    \hat{\B}^\top & \hat{\C}
    \end{bmatrix}$.    
    \State \textbf{Output:} $\hat{\X}$: an approximation of gram matrix $\Xs$.
\end{algorithmic}
\end{algorithm}

We project the estimate $\hat{\X}$ onto the set of rank-$r$ positive semi-definite matrices via truncated eigenvalue decomposition ${\hat\U}_r \hat{\bm{\Lambda}}_r \hat{\U}_r^\top$, where $r=\bar{r}-2$, and set any negative eigenvalues to 0. The estimated coordinates are then computed as $\hat{\P} = \hat{\bm{\Lambda}}_r^{\frac{1}{2}}\hat{\U}_r^\top $. This step yields the $r$-dimensional Euclidean embedding of the nodes centered at the origin. Note that the embedding is unique up to rigid motions.


\section{Numerical experiments}
\subsection{Synthetic data experiments}

We generate $n$ target points $\p_1, \ldots, \p_n$ and $m$ anchor points $\p_{n+1}, \ldots, \p_{m+n}$ in $\mathbb{R}^r$ using the Halton sequence \cite{halton1964algorithm}. The total number of points is fixed at $T = m+n = 500$. The dimensionality $r$ is set to either 2 or 3. Each coordinate is chosen uniformly from the range $[-100, 100]$. The Halton sequence generates points deterministically, ensuring a low-discrepancy distribution that avoids clustering and provides a more even spacing \cite{kocis1997computational}.
These points are concatenated to form a matrix $\P$ of size $r \times T$. 

From $\P$, we construct a squared distance matrix $\D \in \mathbb{R}^{T \times T}$, where $\D$ is calculated using \eqref{eq:P_to_D}. The matrix $\D$ is then partitioned into blocks $\E$, $\F$ and $\G$, as shown in \eqref{eq:D}. We introduce sparse corruptions to the $\F$ block by adding noise to a fraction $\alpha$ of its entries. The corrupted entries are selected uniformly and independently without replacement. The magnitude of the noise is sampled \textit{i.i.d.}~from a uniform distribution over the interval $\left[-\mathbb{E}(|[\F]_{i,j}|), \mathbb{E}(|[\F]_{i,j}|)\right]$.

\begin{figure}[ht]
  \centering
    \vspace{-0.15in}\includegraphics[width=0.32\textwidth]{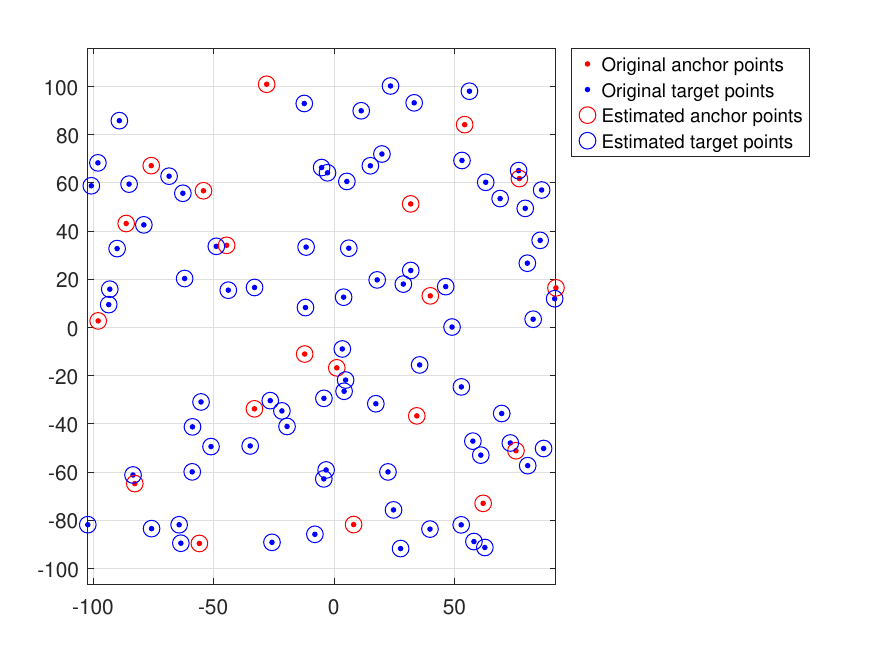}
    \vspace{-0.15in}
    \caption{Visual results for synthetic experiments: The figure illustrates the comparison between original points and estimated points. Here, we set $n=80$, $m=20$, $r=2$ and $\alpha = 0.2$.}
    \vspace{-0.5em}
    \label{fig:fpn}
\end{figure}

To recover the true distances from the corrupted measurements, we employ Structured Robust
EDG (i.e., \Cref{algo:redg}). A truncated eigendecomposition followed by Procrustes analysis \cite{schonemann1966generalized} is used to compute the root mean square error ($\mathrm{RMSE}$), which is defined as:
\begin{equation*}
    \mathrm{RMSE}^2 = {\frac{1}{T} \sum_{i=1}^{T} \left\| \hat{\p}_i - \p_i \right\|_2^2}. 
\end{equation*}
For each combination of $m$ and $\alpha$, we perform 50 trials and report the mean and standard deviation of the $\mathrm{RMSE}$ in \Cref{tab:res_synthetic}.

\begin{table}[ht]
    \centering
    \caption{Synthetic experiment results: Mean $\mathrm{RMSE}$ across 50 trials with standard deviation shown in parentheses. The results are presented for varying numbers of $m$, $\alpha$ and $r$.}
    \label{tab:res_synthetic}
   
    \begin{tabular}{c|ccc|ccc}
        \toprule
         & \multicolumn{3}{c|}{$r=2$} & \multicolumn{3}{c}{$r=3$} \cr
        $m \setminus \alpha$  & 0.10 & 0.20 & 0.30 & 0.10 & 0.20 & 0.30 \cr
        \midrule
        10 & 4.63 & 10.40 & 16.50 & 14.60 & 25.20 & 34.40 \\ 
           & (0.68) & (0.86) & (0.82) & (0.93) & (1.25) & (1.23) \\ \midrule
        20 & 0.28 & 2.20 & 5.65 & 1.05 & 6.43 & 14.40 \\ 
           & (0.55) & (0.81) & (0.95) & (1.26) & (1.09) & (1.10) \\ \midrule
        30 & 0 & 0.30 & 1.90 & 0 & 0.97 & 4.79 \\ 
           & (0) & (0.48) & (0.69) & (0) & (1.10) & (1.07) \\ \midrule
        40 & 0 & 0 & 0.23 & 0 & .25 & 1.25 \\ 
           & (0) & (0) & (0) & (0) & (0.53) & (1.03) \\ \midrule
        50 & 0 & 0 & 0.08 & 0 & 0 & 0.13 \\ 
           & (0) & (0) & (0.28) & (0) & (0) & (0.31) \\ \midrule
        60 & 0 & 0 & 0.02 & 0 & 0 & 0.07 \\ 
           & (0) & (0) & (0.15) & (0) & (0) & (0.23) \\ \bottomrule
    \end{tabular}
\end{table}

\subsection{Protein structure recovery}
We evaluate our algorithm's performance on protein structure data called 1PTQ obtained from the Protein Data Bank \cite{berman2000protein}. The protein consists of 402 atoms in three-dimensional space, providing a real-world test case for our \Cref{algo:redg}. The primary focus of the experiment is to explore how the number of anchor points and corruption fraction influence the accuracy of protein structure reconstruction. 

In this experiment, we utilize a varying number of anchor points $m$, which are selected from the total set of atoms. Although the current methodology uses a uniform sampling technique, we emphasize that in practice, more sophisticated strategies can be applied. These might involve using domain knowledge, such as considering the chemical or spatial proximity of atoms, to inform the anchor selection process.

\begin{figure}[ht]
  \centering
    \includegraphics[width=0.6\linewidth]{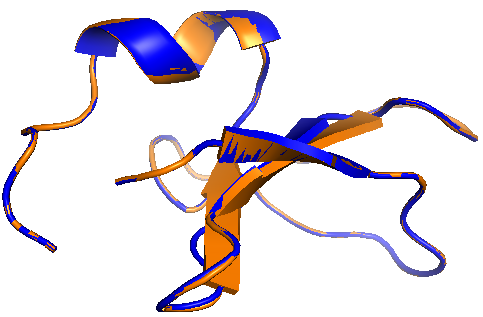}
    \vspace{-0.08in}
    \caption{Target structure 1PTQ (blue) compared to the numerically estimated structure (orange). The visualization corresponds to a single experimental realization with $m = 30$ anchor points and corruption fraction $\alpha = 0.2$ ($\mathrm{RMSE}$ = 0.65).}
    \vspace{-0.5em}
    \label{fig:1ptq}
\end{figure}

To evaluate the accuracy of the reconstructed structure, we compare the aligned estimated protein model to the true structure using $\mathrm{RMSE}$. We conduct the experiment over 50 independent trials. The findings, summarized in \Cref{tab:res_1ptq}, indicate that relatively few anchor points are sufficient to achieve an accurate reconstruction. Notably, even as the corruption fraction $\alpha$  increases, the algorithm demonstrates strong robustness, producing low $\mathrm{RMSE}$ values when an adequate number of anchor points are included. \Cref{fig:1ptq} provides a visual comparison, highlighting the capability of our approach to reliably reconstruct the protein structure, even in the presence of considerable data corruption.

\begin{figure}[ht]
  \centering
    \includegraphics[width=0.6\linewidth]{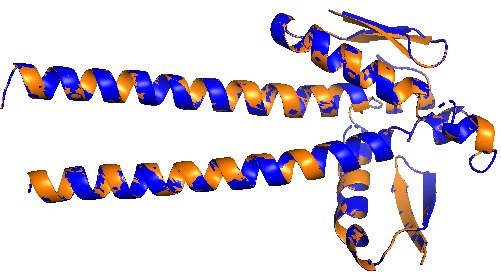}
    \vspace{-0.08in}
    \caption{Target structure 1W2E (blue) compared to the numerically estimated structure (orange). This visualization represents a single experimental realization with $m = 30$ anchor points and a corruption fraction of $\alpha = 0.2$ ($\mathrm{RMSE}$ = $0.71$).}
    \vspace{-0.5em}
    \label{fig:1w2e}
\end{figure}

To assess the scalability of our method, we apply it to the protein 1W2E,  consisting of 2850 atoms. This protein serves as an excellent test case for evaluating our algorithm's performance on large-scale molecular structures. In our experiments, we systematically vary $m$ and evaluate the algorithm’s ability to reconstruct the protein structure with varying levels of $\alpha$. The results, summarized in \Cref{tab:res_1w2e}, show the mean $\mathrm{RMSE}$ across 50 trials at varying levels of $\alpha$ and $m$. \Cref{fig:1w2e} provides a representative reconstruction of the protein 1W2E when $m = 30$ and $\alpha = 0.2$. It highlights the robustness and scalability of our method and further validates the potential of our approach in practical applications.

\begin{table}[hb]
    \centering
    \caption{Protein 1PTQ results: Mean $\mathrm{RMSE}$ (±std) over 50 trials, organized by anchor points ($m$) as a percentage of total atoms (in parentheses) and evaluated across varying $\alpha$.}
    \label{tab:res_1ptq}

    \begin{tabular}{c|cccccc}
        \toprule
        Number of  & \multicolumn{6}{c}{$\alpha$}  \\
        anchors & 0.05 & 0.10 & 0.15 & 0.20 & 0.25 & 0.30 \\
        \midrule
        10 (2.49\%) & 1.45 & 2.34 & 3.01 & 3.87 & 4.43 & 4.99 \\ 
                 & (0.62) & (0.75) & (0.84) & (0.74) & (1.02) & (0.95) \\ \midrule
        20 (4.98\%) & 0.10 & 0.26 & 0.53 & 0.92 & 1.43 & 1.89 \\ 
                 & (0.15) & (0.19) & (0.21) & (0.25) & (0.31) & (0.33) \\ \midrule
        30 (7.46\%) & 0.00 & 0.02 & 0.07 & 0.18 & 0.40 & 0.62 \\ 
                 & (0.03) & (0.05) & (0.09) & (0.14) & (0.15) & (0.15) \\ \midrule
        40 (9.95\%) & 0.00 & 0.00 & 0.00 & 0.02 & 0.09 & 0.24 \\ 
                 & (0.00) & (0.00) & (0.01) & (0.06) & (0.10) & (0.13) \\ \midrule
        50 (12.44\%) & 0.00 & 0.00 & 0.00 & 0.00 & 0.02 & 0.06 \\ 
                 & (0.00) & (0.00) & (0.02) & (0.01) & (0.05) & (0.08) \\ \midrule
        60 (14.99\%) & 0.00 & 0.00 & 0.00 & 0.00 & 0.00 & 0.01 \\ 
                 & (0.00) & (0.00) & (0.00) & (0.00) & (0.00) & (0.03) \\ \bottomrule
    \end{tabular}
\end{table}

\begin{table}[hb]
    \centering
    \caption{Protein 1W2E results: Mean $\mathrm{RMSE}$ (±std) over 50 trials, organized by anchor points ($m$) as a percentage of total atoms (in parentheses) and evaluated across varying $\alpha$.}
    \label{tab:res_1w2e}
   
    \begin{tabular}{c|cccccc}
        \toprule
        Number of  & \multicolumn{6}{c}{$\alpha$}  \\
        anchors & 0.05 & 0.10 & 0.15 & 0.20 & 0.25 & 0.30 \\
        \midrule
        10 (0.35\%) & 4.98 & 7.30 & 9.56 & 10.77 & 11.89 & 13.70 \\ 
           & (2.09) & (2.31) & (2.20) & (3.00) & (2.67) & (2.87) \\ \midrule
        20 (0.70\%) & 0.49 & 1.18 & 2.41 & 3.66 & 4.97 & 6.54 \\ 
           & (0.35) & (0.79) & (1.26) & (1.85) & (1.40) & (1.51) \\ \midrule
        30 (1.05\%) & 0.06 & 0.22 & 0.42 & 0.80 & 1.36 & 2.43 \\ 
           & (0.10) & (0.22) & (0.22) & (0.27) & (0.32) & (1.01) \\ \midrule
        40 (1.40\%) & 0.01 & 0.07 & 0.15 & 0.28 & 0.48 & 0.93 \\ 
           & (0.03) & (0.10) & (0.13) & (0.15) & (0.17) & (0.24) \\ \midrule
        50 (1.75\%) & 0.00 & 0.02 & 0.06 & 0.11 & 0.23 & 0.39 \\ 
           & (0.01) & (0.04) & (0.08) & (0.10) & (0.13) & (0.15) \\ \midrule
        60 (2.10\%) & 0.00 & 0.01 & 0.01 & 0.04 & 0.09 & 0.18 \\ 
           & (0.00) & (0.02) & (0.03) & (0.06) & (0.08) & (0.11) \\ \bottomrule
    \end{tabular}
\end{table}

\section{Conclusion}
In conclusion, this paper introduces a novel algorithm, dubbed Structured Robust EDG, designed for estimating the positions of points with sparse outliers in the distance measurements between anchor and target nodes. The algorithm's effectiveness is demonstrated through empirical evaluations on synthetic sensor localization datasets and molecular experiments, showing that it performs well even with a limited number of anchor nodes and in the presence of substantial sparse noise. Future work will focus on establishing provable error bounds for point reconstruction in Structured Robust
EDG.

\begin{small}
\bibliographystyle{IEEEtran}
\bibliography{IEEEabrv,ref}
\end{small}

\end{document}